\documentclass[11pt, a4paper, logo, two column]{craftjarvis}

\pdftrailerid{redacted}

\makeatletter
\renewcommand\bibentry[1]{\nocite{#1}{\frenchspacing\@nameuse{BR@r@#1\@extra@b@citeb}}}
\makeatother

\usepackage{xcolor}
\definecolor{Gray}{gray}{0.9}
\definecolor{darkgreen}{rgb}{0.0, 0.4, 0.0}   
\definecolor{darkred}{rgb}{0.6, 0.0, 0.0}     
\definecolor{mygreen}{rgb}{0.0, 0.5, 0.0}
\definecolor{myred}{rgb}{0.8, 0.25, 0.33}
\definecolor{myblue}{rgb}{0.19, 0.55, 0.91}
\definecolor{uclablue}{rgb}{0.15, 0.45, 0.68}
\definecolor{zihaoblue}{rgb}{0.25, 0.44, 0.88}
\definecolor{zihaored}{rgb}{0.79, 0.36, 0.27}
\definecolor{boxgreen}{rgb}{0.02, 0.66, 0.02}
\definecolor{boxred}{rgb}{0.66, 0.1, 0.1}
\definecolor{boxblue}{rgb}{0.01, 0.01, 0.73}
\definecolor{langyellow}{RGB}{248, 211, 119}
\definecolor{visgreen}{RGB}{159, 206, 99}
\definecolor{actblue}{RGB}{194, 214, 236}
\definecolor{craftjarviscolor}{RGB}{100, 72, 247}

\usepackage{fancyvrb}

\DefineVerbatimEnvironment{PromptBlock}{Verbatim}{commandchars=\\\{\},fontsize=\small}

\usepackage{listings}
\definecolor{mygray}{gray}{0.4}

\usepackage{times}
\usepackage{graphicx}
\usepackage{amssymb}
\usepackage{url}

\usepackage{ulem}
\usepackage{marvosym}

\usepackage{microtype}
\usepackage{subfigure}
\usepackage{booktabs}

\usepackage{amsmath}
\usepackage{mathtools}
\usepackage{amsthm}

\usepackage{etoc}
\usepackage{algorithm}
\usepackage{algorithmic}
\usepackage{lipsum}
\usepackage{pifont}
\usepackage{wrapfig}
\usepackage{colortbl}
\usepackage{acronym}
\usepackage{multicol}
\usepackage{multirow}
\usepackage{makecell}
\usepackage{enumitem}
\usepackage{tabularx}
\usepackage{colortbl}
\usepackage{array}
\usepackage{xspace}
\usepackage[skip=3pt,font=small]{caption}
\usepackage{xspace}
\usepackage{mdframed}

\usepackage[export]{adjustbox}

\usepackage[utf8]{inputenc} 
\usepackage[T1]{fontenc}    
\usepackage{hyperref}       
\usepackage{amsfonts}       
\usepackage{nicefrac}       

\usepackage{mathtools}
\usepackage{amssymb}
\usepackage{amsthm}

\renewcommand{\paragraph}[1]{\noindent\textbf{#1.}}

\makeatletter
\DeclareRobustCommand\onedot{\futurelet\@let@token\@onedot}
\def\@onedot{\ifx\@let@token.\else.\null\fi\xspace}

\makeatother

\usepackage{xcolor}

\usepackage[most]{tcolorbox}

\newtcolorbox[list inside=prompt,auto counter,number within=section]{prompt}[1][]{
    colbacktitle=black!60,
    coltitle=white,
    fontupper=\footnotesize,
    boxsep=5pt,
    left=0pt,
    right=0pt,
    top=0pt,
    bottom=0pt,
    boxrule=1pt,
    #1,
}

\acrodef{llms}[LLMs]{Large Language Models}
\acrodef{mlms}[MLMs]{Multimodal Language Models}
\acrodef{rag}[RAG]{Retrieval Augmented Generation}
\acrodef{cot}[CoT]{chain-of-thought}

\usepackage[authoryear, sort&compress, round]{natbib} 
\usepackage{bbm}
\title{Training One Model to Master Cross-Level Agentic Actions via Reinforcement Learning}

\reportnumber{} 

\renewcommand{\today}{Dec 2025} 

\author[1]{Kaichen~He}
\author[1]{Zihao~Wang}
\author[1]{Muyao~Li}
\author[2]{Anji~Liu}
\author[1]{Yitao~Liang}

\correspondingauthor{Yitao Liang<yitaol@pku.edu.cn>\\ 
Kaichen He<hkc4623@gmail.com>, Zihao Wang<zhwang@stu.pku.edu.cn>, Muyao Li<2200017405@stu.pku.edu.cn>, Anji Liu<anjiliu@comp.nus.edu.sg>
}

\affil[1]{Peking~University}
\affil[2]{National University of Singapore}
\affil[ \hspace{-0.73ex}]{All authors are affiliated with Team CraftJarvis} 

\begin{abstract}

The paradigm of agentic AI is shifting from engineered complex workflows to post-training native models. However, existing agents are typically confined to static, predefined action spaces—such as exclusively using APIs, GUI events, or robotic commands. This rigidity limits their adaptability in dynamic environments where the optimal granularity of interaction varies contextually.
To bridge this gap, we propose CrossHA, a unified agentic model that masters heterogeneous action spaces and autonomously selects the most effective interface for each step of a trajectory. We introduce a comprehensive training pipeline that integrates cold-start supervised fine-tuning with a Multi-Turn Group Relative Policy Optimization (GRPO) algorithm. This approach enables the agent to learn adaptive action switching—balancing high-level efficiency with low-level precision—without human-specified rules.
Extensive experiments on over 800 tasks in the open-world Minecraft environment demonstrate that CrossHA achieves state-of-the-art performance. By dynamically leveraging the strengths of diverse action spaces, our model significantly outperforms fixed-action baselines, exhibiting superior generalization and efficiency in long-horizon reasoning. All code and models are available at \url{https://github.com/CraftJarvis/OpenHA}.
\end{abstract}

\begin{document}


\maketitle

\section{Introduction}

The field of agentic AI is undergoing a paradigm shift, moving from engineering complex workflows around pre-trained Large Language Models (LLMs)~\citep{llama,chatgpt,qwen3,claude4} or Vision-Language Models (VLMs)~\citep{gpt-4,qwen2vl,gemini,seed15vl} toward developing native agentic models through post-training.
Contemporary native agents are typically characterized by the specific action spaces they master: GUI agents interact via mouse and keyboard events~\citep{operator,uitars}, Deep Research agents utilize API calls~\citep{deepresearchagent,huang2025deep}, Tool-Calling agents integrate with Model Context Protocol (MCP) services~\citep{feng2025retool}, and Vision-Language-Action (VLA) models execute embodied robotic commands~\citep{rt1,rth,rt2}.
However, prior research typically necessitated distinct action space designs for different tasks, often requiring manual selection and definition. To enable interaction within these spaces, researchers had to design specific action translation methods—such as implementing MCP servers~\citep{mcp} for function calls or training learning-based policies for embodied control~\citep{rt-h}.

\begin{figure}
    \centering
    \includegraphics[width=1.0\linewidth]{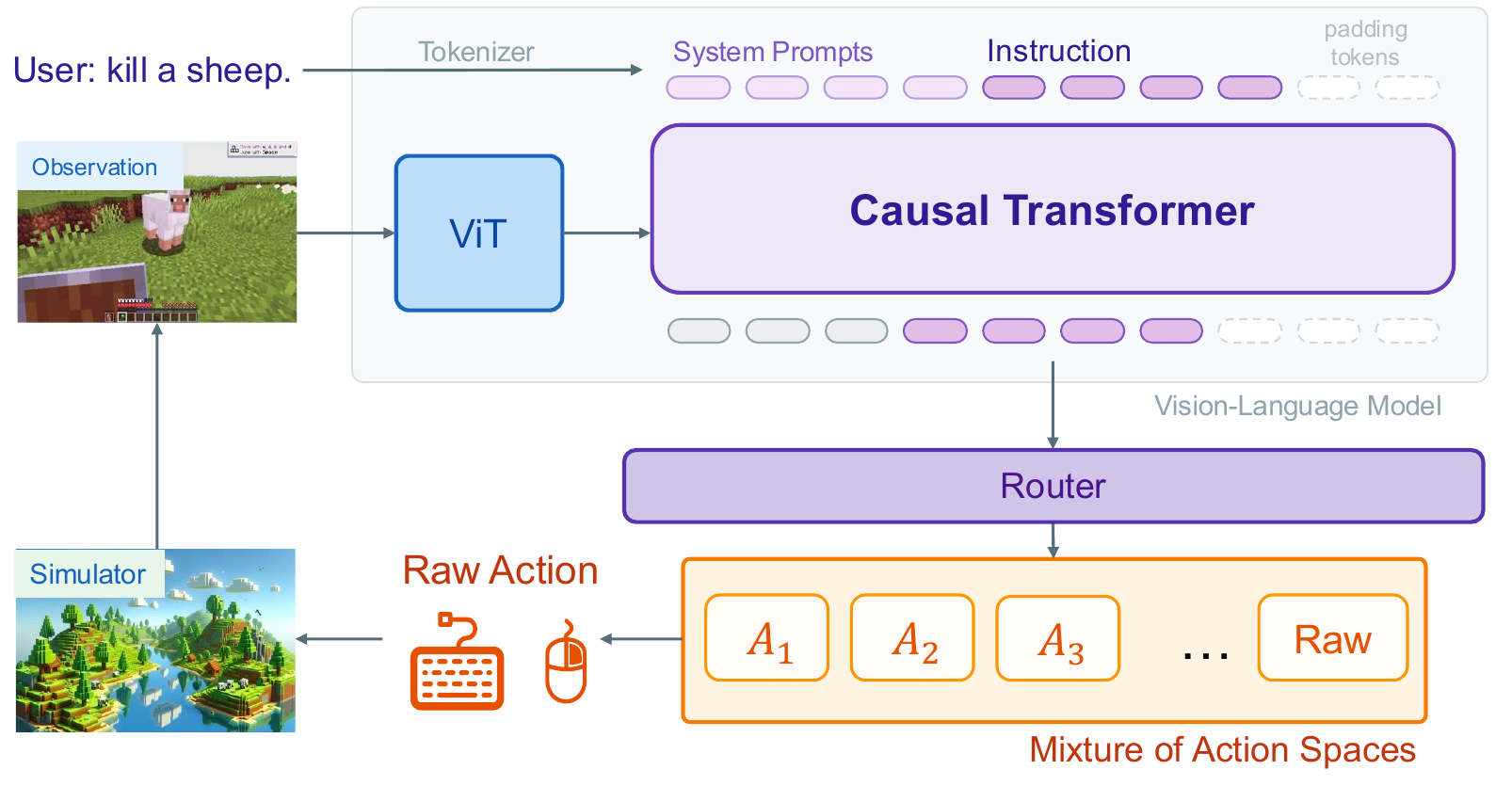}
    \caption{\textbf{The CrossHA Framework}. Unlike prior methods that confine the agent to a fixed action space (e.g., atomic movements) throughout a trajectory, CrossHA dynamically switches across different action spaces to adapt to the context. }
    \label{fig:structure}
\end{figure}

However, relying on statically defined action spaces presents two fundamental challenges.
First, specific action policies or translation layers are often brittle; for instance, an API-based \texttt{read\_url} function may be blocked by CAPTCHA verifications, or a robotic policy may fail to execute a command with perfect precision, thereby capping the agent's success rate~\citep{palme,rt-h}.
Second, the manual assignment of action spaces to tasks restricts the agent's flexibility, preventing it from effectively handling complex scenarios that require multimodal interactions.
Recent works have begun to address this by bridging disjoint spaces to enhance generalization~\citep{uitars2}. Specific approaches include integrating GUI and API-based actions into unified models~\citep{chatgptagent} or merging heterogeneous trajectories during training.

Crucially, we observe that the optimal action space often varies not only across tasks but also within a single task at the step level.
Consider a Deep Research agent: while the majority of information gathering is efficiently handled via search APIs, navigating a specific CAPTCHA-protected webpage may require precise GUI-level manipulation~\citep{deepresearchagent}.
Consequently, a truly generalist agent must possess the ability to dynamically switch between action spaces, ranging from high-level APIs to low-level primitives, to maximize both success rates and efficiency.

To address these limitations, we introduce \textbf{CrossHA}, a native agentic model trained to master multiple heterogeneous action spaces. Unlike traditional approaches that rely on static, human-specified rules, CrossHA autonomously selects the most appropriate action space for each step of a trajectory. We propose a comprehensive training pipeline comprising three stages: cold-start supervised fine-tuning, Single-Turn Reinforcement Learning~(RL), and Multi-Turn RL. By leveraging a Multi-Step Group Relative Policy Optimization (GRPO)~\citep{guo2025deepseek,shao2024deepseekmath} algorithm, the agent learns to optimize its choices at the step level without explicit human intervention. This capability enables the agent to balance trade-offs dynamically; for example, prioritizing high-level actions for efficiency when applicable, while employing fine-grained atomic actions for precise control when necessary.

We validate our approach in the open-world Minecraft environment~\citep{minedojo}~(see \autoref{app:environment_detail} for environment details). Experimental results demonstrate that CrossHA, despite being trained on only 30 tasks, successfully generalizes to over 800 tasks~\citep{mcu}. It autonomously selects optimal action spaces, significantly outperforming baselines confined to fixed action spaces. Furthermore, the agent exhibits emergent behavior by optimizing not only for task success but also for trajectory efficiency, demonstrating robust capabilities for long-horizon reasoning.

Our main contributions are summarized as follows: 1) We propose CrossHA, a unified agentic model capable of mastering heterogeneous action spaces and autonomously selecting the context-appropriate interface without relying on human-defined heuristics. 2) We introduce a comprehensive RL training pipeline utilizing Multi-Turn GRPO, enabling the agent to learn adaptive action switching within a single trajectory to maximize both task success and execution efficiency. 3) We achieve state-of-the-art performance on over 800 tasks in the Minecraft environment, demonstrating that adaptive action-space selection yields superior generalization and robustness compared to static baselines.

\section{Related Works}
\subsection{Agentic Models with Different Actions}
The deployment of VLM and LLM-based agents has expanded rapidly across various domains, ranging from physical environments to digital workspaces, each necessitating distinct interaction paradigms. In embodied settings, approaches such as OpenX~\citep{openx}, RT-H~\citep{rth}, and OpenVLA~\citep{openvla} focus on mapping high-level instructions directly to physical control policies, effectively grounding language into continuous robot trajectories or joint angles~\citep{brohan2023rt,palme}. Transitioning to the digital domain, research on GUI and Web agents, such as Mind2Web and OS-World, centers on visual grounding, where agents learn to manipulate specific UI elements or generate low-level mouse and keyboard events to navigate dynamic screens~\citep{deng2023mind2web,uitars2,li2025screenspot,xu2024aguvis}. In parallel, for logic-intensive tasks such as software engineering, agents like Claude and SWE-agent operate within discrete symbolic spaces, leveraging shell commands or standardized protocols like MCP to seamlessly integrate with external tools and APIs~\citep{claude4,yang2024sweagent,feng2025retool,mcp,wang2024rat}. Similarly, many VLM-based game agents interact with video games directly via mouse and keyboard commands~\citep{wang2025game,openha,jarvisvla}.

\subsection{Generalist Agent with Multi-Action Space}
The static design of action spaces creates limitations when facing dynamic environments~\citep{openha}. A common solution is to leverage the powerful in-context learning capabilities of foundation models~\citep{chatgpt, claude4, gemini} to orchestrate workflows that specify transitions between action spaces for specific scenarios. For example, some computer-use agents~\citep{operator, coact} integrate tool usage~\citep{deepresearchagent} with GUI actions to perform complex tasks: utilizing GUI actions for frontend interface validation while employing MCP when information retrieval is required. However, this approach often necessitates the design of complex, brittle pipelines.

A more unified approach involves fine-tuning agents across multiple action spaces to create a single generalist model. Prior explorations have demonstrated promising results in various domains, such as embodied agents~\citep{onetwovla, univla}, game agents~\citep{wang2025game, openha, jarvis1}, and computer use~\citep{uitars2}. Experiments have demonstrated~\citep{OSWorld_MCP, MCP_world} that agents mastering distinct action spaces exhibit strong generalization capabilities and enhanced performance within specific domains.

However, previous efforts have often overlooked the potential of learning from experience to optimize these transitions. Our approach builds on prior work by advancing this paradigm, leveraging reinforcement learning to enable the model to autonomously adapt to the most appropriate action space at any given step.

\section{Method}
\label{sec:method}

\begin{figure*}
    \centering
    \includegraphics[width=1.0\linewidth]{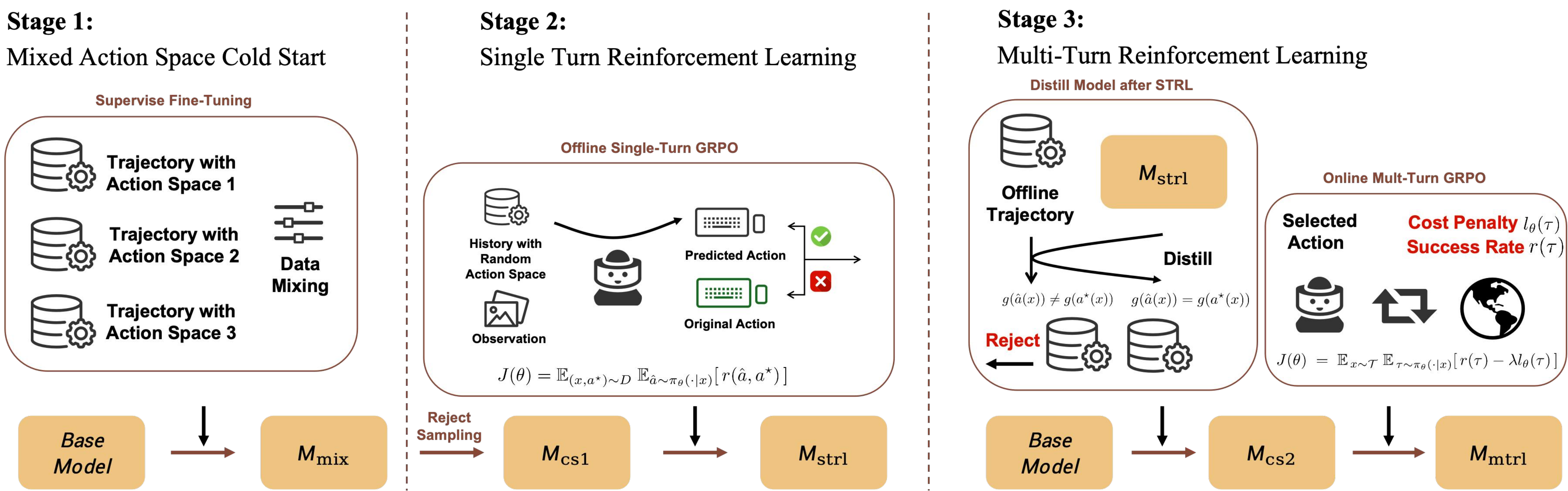}
    \caption{\textbf{Overview of the CrossHA Training Pipeline.} The pipeline comprises three distinct stages: Cold-Start Supervised Fine-Tuning (SFT), Single-Turn Reinforcement Learning (STRL), and Multi-Turn Reinforcement Learning (MTRL). In the first stage, the model learns to decode actions from a heterogeneous action space using a balanced dataset. During STRL, the model is fine-tuned to autonomously select the appropriate action space based on the immediate task context. Finally, in the MTRL stage, the policy is further optimized to balance task success rate with execution efficiency over long horizons. This progressive pipeline ensures CrossHA effectively adapts its action granularity across a wide range of tasks.}
    \label{fig:pipeline}
\end{figure*}

In this section, we introduce our approach for building an agent capable of autonomously selecting the most appropriate action space based on the task context.

We first formulate the problem. Consider an embodied control task modeled as a Markov Decision Process (MDP) with a state space $\mathcal{S}$ and a composite action space:
\[
\mathcal{A}=\bigcup_{x=1}^{N}\mathcal{A}_x .
\]
Each subspace $\mathcal{A}_x$ corresponds to a distinct class of actions, ranging from low-level motor controls to high-level motion primitives. Each subspace is associated with a specific interface or controller $C_x$ that executes the abstract action $a_t\in\mathcal{A}_x$ within the environment.

At each timestep, the agent must determine \textbf{both} the optimal action space and the specific action content, balancing the immediate reward with execution efficiency. This leads to the following objective:
\[
J = \mathbb{E}\left[\sum_t \left(r_t - \lambda_x\, \text{cost}(a_t)\right)\right],
\]
where the term $\lambda_x\, \text{cost}(a_t)$ penalizes the computational or operational cost associated with different action granularities (e.g., token length or execution time).

The central challenge lies in the fact that different tasks, and even different phases within the same task, naturally favor different action spaces. Naive training on a unified space often fails to develop reliable, context-aware selection behavior. To address this, we introduce a phased training curriculum that progressively builds this capability. Beginning with supervised initialization, proceeding to strengthen local preferences through single-turn optimization, and culminating in long-horizon reasoning via multi-turn reinforcement learning, our framework equips the model with robust, dynamic action-space selection capabilities.

The overall training pipeline is illustrated in \autoref{fig:pipeline}. We detail each stage in the following subsections.

\subsection{Stage-1: Mixed-Space SFT}

In the initial stage, we establish the model's foundational capability to execute actions across heterogeneous spaces. We employ Supervised Fine-Tuning (SFT) on a balanced dataset comprising trajectories drawn from multiple action subspaces. By unifying data from these diverse modalities, the model learns to decode and generate valid actions within a composite action space. The primary objective of this stage is to construct a robust base model capable of grounding instructions into executable actions from diverse sources, prior to introducing complex decision-making tasks.

It is important to note that the resulting model, denoted as $M_{mix}$, does not yet autonomously select the optimal action space. Instead, it focuses on learning the syntax and semantics of multiple action types from the mixed data representation. The key challenge addressed here is enabling the model to interpret and generate actions for different spaces without modal interference, thereby laying a solid foundation for the adaptive decision-making mechanisms introduced in subsequent stages.

\subsection{Stage-2: Single-Turn RL}

In this stage, we empower the model to autonomously select the most appropriate action space using \textbf{Single-Turn Reinforcement Learning (STRL)}.

\paragraph{Warm-up: Diversity-Enhanced SFT.}
Standard datasets typically lack explicit annotations indicating which action space is optimal for a given context. To bridge this gap, we first perform a warm-up phase to expose the model to multiple viable paths. We construct prompts that encourage the model to generate candidate actions across all available action spaces. These candidates are filtered via a \textbf{rejection sampling} approach: only actions that successfully execute the task (verified by the environment or a parser) are retained as ground-truth annotations.
We then fine-tune the model on this rebalanced, multi-space dataset. The resulting model, denoted as $M_{cs1}$, acquires the capability to generate valid actions in various formats. However, empirical observation reveals that while $M_{cs1}$ can operate across spaces, its selection strategy remains stochastic rather than strategic—it does not yet deliberately optimize the interface choice for task efficiency or robustness.

\paragraph{Optimization via GRPO.}
To transition from stochastic capability to strategic selection, we employ Group Relative Policy Optimization (GRPO)~\cite{shao2024deepseekmath}, a reinforcement learning algorithm distinguished by its stability and resource efficiency. Unlike traditional methods like PPO that rely on a separate value function network for advantage estimation, GRPO utilizes the group statistics of multiple sampled outputs to estimate the baseline, thereby simplifying the optimization process.

Formally, for each query $q$, a group of outputs $\{o_1, o_2, \dots, o_G\}$ is sampled from the current policy $\pi_{\text{old}}$. The policy $\pi_\theta$ is then optimized via the following objective:

\begin{align}
J_{\text{GRPO}}(\theta) =& \mathbb{E}_{q \sim P(Q), \{o_i\}_{i=1}^G \sim \pi_{\text{old}}} \bigg[ \frac{1}{G} \sum_{i=1}^{G} \frac{1}{|o_i|} \sum_{t=1}^{|o_i|} \nonumber \\ &\min \bigg( \nonumber
\rho_{i,t} \hat{A}_{i,t}, \text{clip}\left(\rho_{i,t}, 1 - \epsilon, 1 + \epsilon \right) \hat{A}_{i,t} \bigg)\\ & - \beta D_{\text{KL}} [ \pi_\theta || \pi_{\text{ref}} ] \bigg]
\label{eq:st_grpo}
\end{align}
where $\rho_{i,t} = \frac{\pi_\theta(o_i | q, o_{<t})}{\pi_{\text{old}}(o_i | q, o_{<t})}$ is the probability ratio. Crucially, $\hat{A}_{i,t}$ represents the advantage estimate, which is calculated based on the relative performance of output $o_i$ compared to the group mean, rather than a learned value function. This encourages the model to favor actions that outperform their peers within the same sampled group.

Since this stage focuses on immediate, one-step decision-making (viewing the task as a single-turn problem to optimize action space selection), we term it the STRL stage. To endow the cold-start model $M_{cs1}$ with \textbf{action-space autonomy}, we cast each example in the dataset $D$ as a one-step decision problem. Let $\{\mathcal{A}_k\}_{k=1}^K$ denote the set of string-level action spaces and $\mathcal{A} = \bigcup_{k} \mathcal{A}_k$. We utilize a deterministic parser $g:\mathcal{A}\to\mathcal{R}$ that maps an action string to its canonical raw representation in $\mathcal{R}$. The reward function is defined as:
\begin{equation}
\label{eq:strl_reward}
r(\hat{a}, a^\star) = \mathbbm{1}\left\{ g(\hat{a}) = g(a^\star) \right\},
\end{equation}
This reward is \textbf{action-space agnostic}: credit is granted whenever the parsed raw action matches the ground truth, regardless of the surface form (action space) used to generate $\hat{a}$. The STRL objective is to maximize the expected reward:
\begin{equation}
\label{eq:strl_objective}
J(\theta) = \mathbb{E}_{(x, a^\star) \sim D} \; \mathbb{E}_{\hat{a} \sim \pi_\theta(\cdot|x)} \left[ r(\hat{a}, a^\star) \right].
\end{equation}
Through this process, the model learns to ignore prior biases and autonomously select whichever action space most reliably yields the correct raw action for a given input. We denote the resulting model as $M_{strl}$.

\subsection{Stage-3: Multi-Turn RL}

Although \(M_{strl}\) achieves high accuracy in single-step action prediction, its probability distribution tends to be excessively peaked around specific action spaces, which limits exploration and hinders long-horizon task success. To explicitly cultivate robust, trajectory-level action-space selection, we adopt \textbf{Multi-Turn Reinforcement Learning (MTRL)}. In this stage, the optimization signal is derived from the episodic success rate, guiding the model to refine its policy for maximizing overall task completion rather than just immediate correctness.

\paragraph{Initialization via Self-Training ($M_{cs2}$).}
To accelerate training, we first distill the action-space preferences of \(M_{strl}\) back into a supervised format. We perform inference over the initial dataset \(D\) using \(M_{strl}\).
For each example \((x,a^\star)\), let \(\hat a\) denote the model's prediction and let \(g(\cdot)\) map an action string to its canonical raw representation.
We construct a relabeled dataset \(D_{strl}\) via the following rule:
\begin{equation}
\label{eq:strl_relabel}
a'(x) \;=\;
\begin{cases}
\hat a(x), & \text{if } g(\hat a(x)) = g(a^\star(x)),\\[2pt]
a^\star(x), & \text{otherwise}.
\end{cases}
\end{equation}
Essentially, if the predicted action is semantically consistent with the ground truth (as verified by \(g\)), we replace the original label with the action space chosen by \(M_{strl}\); otherwise, we retain the original label.
This relabeling preserves the raw-action semantics while aligning the surface-level action space with the model's learned preferences.
We then fine-tune the base model on this relabeled dataset \(D_{strl}\), following the same protocol as the previous SFT stage.
We denote the resulting model as \(M_{cs2}\). Since the ground truth in \( D_{\text{strl}} \) reflects the optimized choices of \( M_{\text{strl}} \), \( M_{\text{cs2}} \) initializes the MTRL stage with a strong prior for appropriate action-space selection.

\paragraph{Trajectory Optimization.}
We subsequently fine-tune \(M_{cs2}\) using multi-turn RL on a curated task set \(\mathcal{T}\).
Given an instruction \(x \in \mathcal{T}\), the policy \(\pi_\theta(\cdot \mid x)\) interacts with the environment to generate a trajectory \(\tau\).
We utilize a binary episodic reward:
\begin{equation}
\label{eq:mtrl-reward}
r(\tau) = \mathbbm{1}\{\text{success}(\tau)\}
\end{equation}
The objective is to maximize the task success rate while penalizing excessive generation costs:
\begin{equation}
\label{eq:mtrl-objective}
J(\theta) \;=\;
\mathbb{E}_{\,x \sim \mathcal{T}}\;
\mathbb{E}_{\,\tau \sim \pi_\theta(\cdot \mid x)}\!
\left[\, r(\tau) - \lambda l_\theta(\tau) \,\right].
\end{equation}
where $l_\theta(\tau)$ denotes the total number of tokens produced by the model throughout the trajectory. This penalty term encourages the model to prefer concise action spaces (e.g., high-level APIs) over verbose ones (e.g., raw primitive commands) when both yield success. A detailed discussion of the specific GRPO implementation for this stage is provided in \autoref{app:rl_algorithm}.

By optimizing \autoref{eq:mtrl-objective}, the model improves its capacity for dynamic, context-aware action space adaptation over long horizons. We denote the final model as \( M_{\text{mtrl}} \), which serves as our final \textbf{CrossHA}.

\section{Experiments}

We present a comprehensive evaluation to validate our framework, focusing on how heterogeneous action spaces affect performance in open-ended environments. 
We describe the experimental setup, including training protocols, benchmarks, and evaluation metrics, and address the following questions:

\textbf{Q1: Action Space Sensitivity.} How does the choice of a fixed action space impact performance, and does the optimal space vary across task categories?

\textbf{Q2: Benefits of Dynamic Switching.} Does dynamic action-space selection outperform static baselines?

\textbf{Q3: Generalization and Robustness.} Does the model generalize to unseen tasks after multi-turn RL fine-tuning on a limited set of tasks?
\begin{table*}[]
    \centering
    \caption{
        The evaluation results of Minecraft agents across more than 800 tasks are presented.For each task category, we report three metrics: the success rate of a representative task (indicated by its icon), the percentage of tasks that the agent succeeds at least once (FT), and the average success rate across all tasks in the category (ASR) with standard deviation. Results highlighted in {\color{blue}{blue}} correspond to the second-best performances, while those in {\color{red}{red}} represent the state-of-the-art performance for each metric across all agents.
    }
    \label{tab:main_results}
    \resizebox{\textwidth}{!}{%
    \renewcommand\arraystretch{1.4}
    \begin{tabular}{@{}lcccccccccccccc@{}}
    \toprule
     & \multicolumn{3}{c}{Mine Blocks} &  & \multicolumn{3}{c}{Kill Entities} &  & \multicolumn{3}{c}{Craft Items} & & \multicolumn{2}{c}{All Tasks}\\ \cmidrule(l){2-4} \cmidrule(l){6-8} \cmidrule(l){10-12} \cmidrule(l){14-15}
    \multirow{-2}{*}{Method} & \includegraphics[scale=0.04,valign=c]{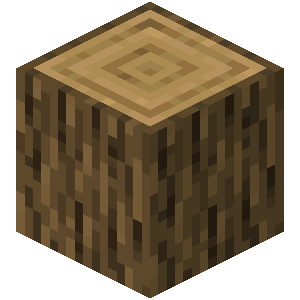} & \textit{FT} $\uparrow$ & \textit{ASR} $\uparrow$ &  & \includegraphics[scale=0.16,valign=c]{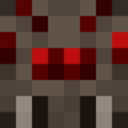} & \textit{FT} $\uparrow$ & \textit{ASR} $\uparrow$ &  & \includegraphics[scale=0.08,valign=c]{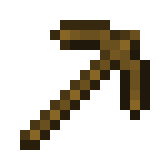} & \textit{FT} $\uparrow$ & \textit{ASR} $\uparrow$ &  &\textit{FT} $\uparrow$ & \textit{ASR} $\uparrow$ \\ \midrule
    \multicolumn{15}{l}{\cellcolor[HTML]{E1EAFF}{\textbf{\textit{Instruction-Conditioned Policies}}}}\\
    VPT~\citep{vpt} & 20.0 & 30.7 & $6.0^{\pm11.4}$ &  & 10.0 & 24.6 & $3.6^{\pm7.7}$ &  & 0.0 & 6.7 & $0.8^{\pm3.3}$& & 20.7 &$3.5^{\pm8.4}$\\
    STEVE-I~\citep{steve1} & 50.0 & 29.4 & $8.0^{\pm17.0}$ &  & 0.0 & 14.7 & $3.9^{\pm12.0}$ &  & 0.0 & 16.4 & $3.2^{\pm8.4}$& & 20.2 & $5.0^{\pm12.4}$\\
    ROCKET-1~\citep{rocket1} & 60.0 & 57.5 & $18.9^{\pm24.3}$ &  & 60.0 & \textbf{\color{blue}{63.9}} & $27.9^{\pm29.3}$ &  & 0.0 & 0.0 & $0.0^{\pm{0.0}}$& & 45.5 &$15.6^{\pm{24.9}}$\\
    JARVIS-VLA~\citep{jarvisvla} & 55.0 & 55.3 & $30.0^{\pm35.4}$ &  & 60.0 & 61.9 & $18.5^{\pm22.7}$ &  & 40.0 & \textbf{\color{red}{74.3}} & $25.1^{\pm23.9}$& & \textbf{\color{red}{63.8}} &$24.5^{\pm{28.4}}$\\ 
    \midrule
    \multicolumn{15}{l}{\cellcolor[HTML]{F0FBEF}{\textbf{\textit{VLM-based Agents}}}} \\
    TextHA~\citep{openha} & 60.0 & 36.3 & $27.2^{\pm38.2}$ 
    &  & 0.0 & 19.2 & $8.7^{\pm23.7}$ 
    &  & 50.0 & 43.9 & $26.0^{\pm35.2}$ & & 33.1 &$20.6^{\pm{34.0}}$\\
    GroundingHA~\citep{rocket1} & \textbf{\color{blue}{90.0}} & \textbf{\color{blue}{61.0}} & $37.1^{\pm38.5}$
        &  & 50.0 & \textbf{\color{red}{90.1}} & $26.5^{\pm23.4}$ 
        &  & 15.0 & 27.5 & $6.7^{\pm10.8}$ & & 59.5 &$23.4^{\pm{29.6}}$\\
    UI-TARS-1.5~\citep{uitars15} & - & - & $42.1^{\pm20.4}$
        &  & - & - & $31.0^{\pm16.4}$ 
        &  & 0 & 0 & $36.7^{\pm17.2}$ & & - &$33.8$\\
    MotionHA~\citep{rth} & 70.0 & 51.0 & $27.4^{\pm35.2}$ 
        &  & 20.0 & 29.5 & $4.3^{\pm10.8}$ 
        &  & 0.0 & 0.0 & $0.0^{\pm0.0}$ & & 29.8 &$10.6^{\pm{24.4}}$\\
    LanguageHA~\citep{palme} & 60.0 & 31.3 & $11.3^{\pm14.5}$ 
    &  & 0.0 & 12.8 & $6.5^{\pm9.3}$ 
    &  & 5.0 & 19.3 & $6.3^{\pm9.2}$ & & 21.1 &$8.0^{\pm{11.5}}$\\
    LatentHA~\citep{omnijarvis} & 70.0 & 54.2 & $24.4^{\pm31.1}$ 
        &  & 50.0 & 24.6 & $8.5^{\pm17.9}$ 
        &  & 0.0 & 19.1 & $3.0^{\pm7.5}$ & & 32.6 &$12.0^{\pm{23.0}}$\\
    OpenHA~\citep{openha} & 80.0 & \textbf{\color{red}{67.3}} & $30.1^{\pm13.9}$ 
        &  & \textbf{\color{red}{70.0}} & 62.6 & $32.5^{\pm9.2}$
        &  & \textbf{\color{blue}{80.0}} & 58.8 & $31.9^{\pm13.7}$ & & \textbf{\color{blue}{62.8}} &$31.5^{\pm{12.5}}$\\
    Game-TARS~\citep{wang2025game} & - & - & \textbf{\color{red}{$50.14^{\pm20.7}$}} 
        &  & - & - & \textbf{\color{blue}{$38.1^{\pm24.6}$}}
        &  & - & - & $39.1^{\pm27.5}$ & & - &\textbf{\color{blue}{$42.2$}}\\
    \midrule
    \multicolumn{15}{l}{\cellcolor[HTML]{FEF1F1}{\textbf{\textit{Ours}}}} \\
    CrossHA(w/o STRL stage) 
    & 63.6 & 35.1 & $39.0^{\pm46.5}$ 
    & & 28.6& 43.9 & $27.7^{\pm43.9}$ 
    & & 57.1 & 55.7 & \textbf{\color{blue}{$58.0^{\pm48.4}$}} & & 44.9 &$41.6^{\pm{47.9}}$\\
    CrossHA & \textbf{\color{red}{94.7}} & 45.2 & \textbf{\color{blue}{$40.0^{\pm48.3}$}}
    & & \textbf{\color{blue}{66.6}} & 58.1 & \textbf{\color{red}{$45.1^{\pm43.5}$}} 
    & & \textbf{\color{red}{83.3}} & \textbf{\color{blue}{72.7}} & \textbf{\color{red}{$78.8^{\pm41.0}$}} & & 58.7 &\textbf{\color{red}{$54.6^{\pm47.6}$}}\\
    \bottomrule
    \end{tabular}%
}
\vspace{-0.1 in}
\end{table*}

\subsection{Experimental Setup}

\paragraph{Benchmark.}
We utilize Minecraft (version 1.16.5) as our primary evaluation environment~\citep{minerl}. The agent’s observation space consists exclusively of first-person RGB images with a resolution of $360 \times 640 \times 3$. The action space employs a discrete, human-like interface, comprising mouse movements, clicks, and keyboard commands. Further details regarding the observation and action spaces are provided in \autoref{app:environment_detail}.

To assess our agent, we employ the OpenHA benchmark suite~\citep{openha}, which features over 800 tasks. All tasks are manually designed and verified, ensuring both richness and feasibility. We categorize these tasks into three groups based on the primary skills required:
1) \textbf{Mine Blocks}: Tasks involving navigation and physical interaction within the environment (e.g., locating and chopping down a specific type of tree);
2) \textbf{Craft Items}: Tasks requiring complex interactions with graphical user interfaces (GUIs), such as crafting at a table or smelting ores in a furnace;
3) \textbf{Kill Entities}: Tasks focusing on survival and combat, requiring the agent to engage with dynamic mobs.

In our evaluation protocol, we adopt two primary metrics: \textbf{Finished Tasks (FT)} and \textbf{Average Success Rate (ASR)}.
FT measures the proportion of distinct tasks within a category for which the agent achieves at least one successful completion, reflecting the agent’s \textit{task coverage} and generalization range.
ASR quantifies the mean success rate across all tasks in the category, capturing \textit{reliability} by accounting for the consistency of the agent's performance on each specific task.
\paragraph{Training Setup.}
1) \textbf{Model Initialization}. We initialize our training with Qwen2-VL-7B-Instruct~\citep{qwen2vl}, a pre-trained vision-language model demonstrating strong multimodal understanding. We fine-tune this model using diverse Minecraft-specific VQA and captioning datasets~\citep{jarvisvla}, yielding the base model $M_{base}$.
2) \textbf{Dataset Construction}. We generate grounding and motion ground-truth annotations by integrating two systems: a SAM-based~\citep{sam2} grounding pipeline and a motion-generation module built upon a fine-tuned MineCLIP~\citep{minedojo} model. These systems annotate both VPT data and a subset of contractor-collected trajectories. Since suitable grounding or motion annotations cannot be generated for every trajectory, the resulting action space distribution is imbalanced. These annotated data are integrated to form the mixed-action-space dataset $D_{mix}$.

We then process $M_{base}$ and $D_{mix}$ through the pipeline described in \autoref{sec:method}. During the MTRL stage, we select 10 tasks from each of the three primary OpenHA categories: \texttt{craft\_item}, \texttt{kill\_entity}, and \texttt{mine\_block}. This results in a total of 30 training tasks. For online RL, we train using GRPO for over 80 iterations, with each iteration involving over 6,400 environment interactions. This extensive training regime allows the model's performance to converge to a stable, high-level policy.
\paragraph{Baselines.}
Our analysis compares CrossHA against variants of the base model $M_{base}$ fine-tuned (via either SFT or RL) on datasets restricted to single-type action spaces. We categorize these baselines based on their action space formats: LanguageHA, GroundingHA, MotionHA, RawHA, and LatentHA.
We also include several representative LLM-based agents, such as Jarvis-VLA~\citep{jarvisvla}, and hierarchical agents fine-tuned with SFT on fixed action spaces. 
In addition, we consider recent unified agent frameworks such as OpenHA~\citep{openha}, which integrate multiple action spaces within a hierarchical agent architecture and enable flexible switching during execution. 
However, such switching behavior is not explicitly optimized as a learnable decision variable, in contrast to CrossHA, which formulates action-space selection as a policy learning problem and optimizes it via reinforcement learning.
For broader context, we further compare against specialized policies trained on the VPT dataset, including the original VPT~\citep{vpt}, ROCKET-1~\citep{rocket1}, and STEVE-1~\citep{steve1}, as well as Game-TARS~\citep{wang2025game}, a large-scale pretrained game agent that leverages multimodal pretraining and hierarchical control for general gameplay.

\subsection{Main Results}
As presented in \autoref{tab:main_results}, CrossHA achieves state-of-the-art performance across all three major task categories, substantially outperforming existing instruction-conditioned policies and hierarchical agents. Our analysis highlights three key observations:

\paragraph{Clear Advantages Over Single-Action-Space Agents.}
Agents confined to a single action space exhibit strong but narrow specialization, leading to pronounced performance asymmetries across task types. GroundingHA excels at \textit{Kill Entity} tasks (FT: 90.1\%), MotionHA performs better on \textit{Mine Block} tasks, and RawHA shows advantages in \textit{Craft Item} tasks due to fine-grained control. However, none of the single-action-space baselines achieve uniformly strong results, underscoring the limitations of rigid action-space-specific designs.

\paragraph{RL Facilitates Robust OOD Generalization.}
Comparing SFT-only agents with their RL-enhanced counterparts reveals consistent gains in both ASR and FT. Although RL fine-tuning is conducted on only 30 tasks, the improvements generalize to over 800 evaluation tasks, especially in fine-grained control domains such as \textit{Craft Item}. This confirms the importance of RL on top of large pre-trained VLA models for scalable OOD generalization.

\paragraph{CrossHA Achieves Balanced, Omni-Category Performance.}
CrossHA achieves the best or second-best results in nearly every metric across all categories. It balances proficiency across \textit{Mine Blocks}, \textit{Kill Entities}, and \textit{Craft Items}, while attaining high peak success rates such as 94.7\% in Mine Blocks and 83.3\% in Craft Items. These results strongly suggest that CrossHA learns to coordinate and select the most suitable action space based on task context.

\begin{figure}
    \centering
    \includegraphics[width=1.0\linewidth]{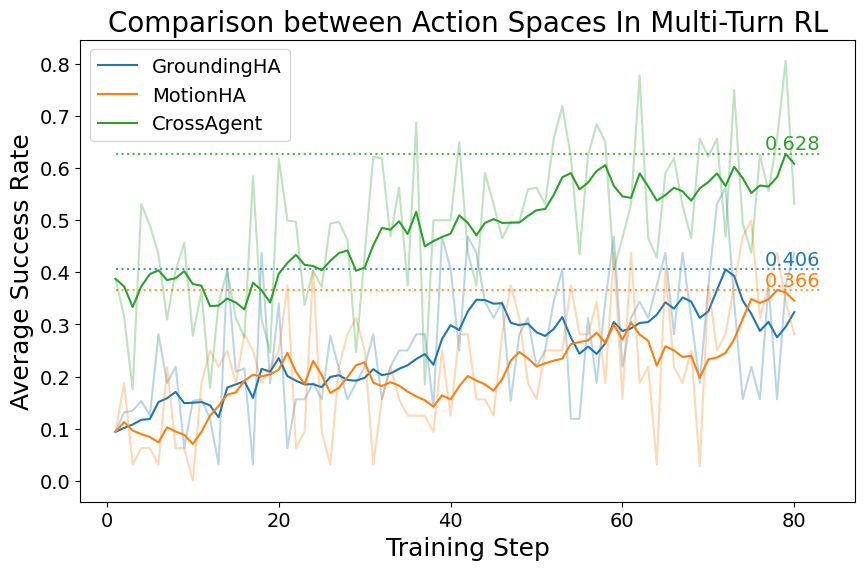}
    \caption{
    \textbf{Performance Comparison Across Action Spaces.}
    The heterogeneous action space of CrossHA enables superior data efficiency and higher asymptotic performance during multi-turn reinforcement learning, compared to single-space baselines.
    }
    \label{fig:ablation_coa}
\end{figure}

\begin{figure}
    \centering
    \includegraphics[width=1.0\linewidth]{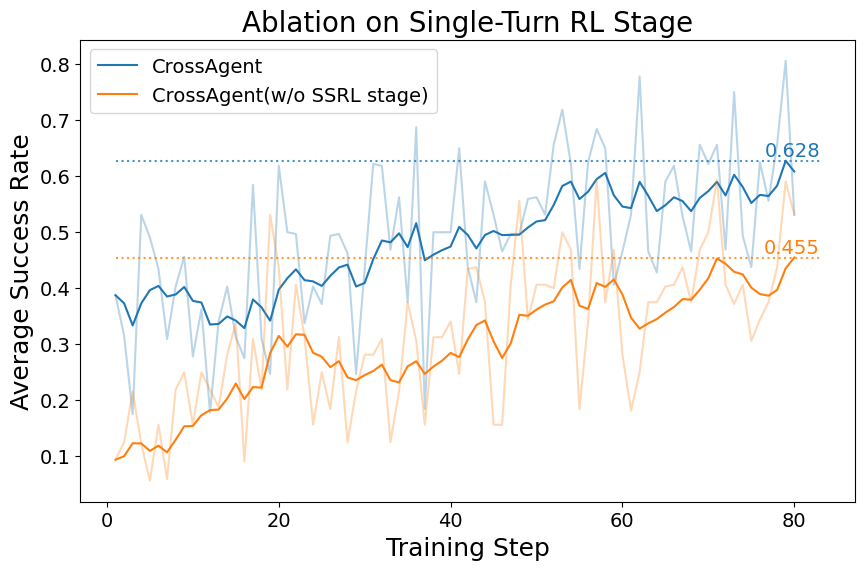}
    \caption{\textbf{Effect of the Single-Turn RL (STRL) Stage.}
    Training curves comparing CrossHA with and without the STRL phase. The inclusion of STRL significantly enhances training efficiency and accelerates convergence in the subsequent MTRL stage, despite its low computational cost.}
    \label{fig:ablation_strl}
\end{figure}

\subsection{Ablation Studies}

\begin{table*}[t]
    \centering
    \caption{
        Evaluation results of RL agents on In-Distribution (ID) and Out-of-Distribution (OOD) tasks. 
        We report the success rate and standard deviation. 
        \textbf{Red} indicates the best performance, and \textbf{Blue} indicates the second best.
    }
    \label{tab:generalization_ablations}
    \resizebox{\textwidth}{!}{%
    \begin{tabular}{@{}lcccccccc@{}}
    \toprule
    \multirow{3}{*}{Method} & \multicolumn{4}{c}{\textbf{In-Distribution Evaluation}} & \multicolumn{4}{c}{\textbf{Out-of-Distribution Evaluation}} \\
    \cmidrule(lr){2-5} \cmidrule(l){6-9}
      & Mine Blocks & Kill Entities & Craft Items & All Tasks & Mine Blocks & Kill Entities & Craft Items & All Tasks \\ 
    \midrule
    
    RawHA-RL 
    & $65.2^{\pm32.1}$ 
    & $41.6^{\pm21.8}$ 
    & \color{red}{$96.2^{\pm3.8}$} 
    & \color{red}{$70.1^{\pm33.6}$}
    & $28.9^{\pm44.2}$
    & \color{blue}{$28.4^{\pm32.3}$}
    & \color{blue}{$69.8^{\pm44.4}$}
    & \color{blue}{$42.4^{\pm45.1}$} \\

    GroundingHA-RL 
    & $46.8^{\pm35.6}$
    & \color{blue}{$48.8^{\pm34.5}$}
    & $69.3^{\pm28.4}$
    & $52.6^{\pm{31.5}}$
    & $27.2^{\pm42.4}$
    & \color{red}{$33.9^{\pm35.9}$}
    & $57.2^{\pm48.1}$
    & $39.4^{\pm{44.3}}$ \\

    MotionHA-RL 
    & \color{red}{$78.6^{\pm24.6}$}
    & $44.0^{\pm25.2}$
    & $63.0^{\pm20.7}$
    & $61.9^{\pm{27.5}}$
    & \color{red}{$41.5^{\pm45.5}$}
    & $26.8^{\pm26.7}$
    & $49.0^{\pm47.3}$
    & $39.1^{\pm{42.0}}$ \\

    
    CrossHA(w/o STRL) 
    & $63.3^{\pm28.9}$
    & $24.0^{\pm26.4}$
    & $76.1^{\pm27.2}$
    & $54.5^{\pm{35.3}}$
    & $39.0^{\pm46.5}$
    & $22.2^{\pm42.6}$
    & $58.0^{\pm48.4}$
    & $39.7^{\pm{48.1}}$ \\

    CrossHA 
    & \color{blue}{$70.7^{\pm33.3}$}
    & \color{red}{$52.1^{\pm22.9}$}
    & \color{blue}{$83.7^{\pm25.5}$}
    & \color{blue}{$68.8^{\pm30.5}$}
    & \color{blue}{$40.0^{\pm48.3}$}
    & \color{blue}{$28.4^{\pm33.2}$}
    & \color{red}{$78.8^{\pm41.0}$}
    & \color{red}{$49.1^{\pm46.6}$} \\
    
    \bottomrule
    \end{tabular}%
    }
\end{table*}

\paragraph{Impact of Mixed Action Spaces.}
To evaluate the benefit of a unified heterogeneous action space, we compare CrossHA with single-space baselines, \textit{GroundingHA} and \textit{MotionHA}. These baselines initialize MTRL with only the grounding-space or motion-space subset of $D_{strl}$, while CrossHA uses the full mixed dataset. All models undergo the same 200-step SFT cold start before MTRL.

As shown in \autoref{fig:ablation_coa}, CrossHA converges faster and reaches higher asymptotic performance than single-space baselines, indicating that dynamic action-space selection provides greater flexibility and improves RL data efficiency.

\paragraph{Necessity of the STRL Stage.}
We further study the role of STRL. For the variant \textit{without} STRL, we skip Stage 2 and initialize MTRL with the original balanced dataset $D_{bal}$ instead of the STRL-relabeled dataset $D_{strl}$.

As shown in \autoref{fig:ablation_strl}, STRL improves both sample efficiency and final performance by providing an effective warm start for action-space selection. Results in \autoref{tab:main_results} further show that CrossHA with STRL achieves higher success rates across a broader range of tasks.

\subsection{Generalization Evaluation}

While \autoref{tab:main_results} shows strong OOD results, we further compare In-Distribution (ID) training tasks and OOD evaluation tasks in \autoref{tab:generalization_ablations} to examine whether RL fine-tuning harms generalization.

\paragraph{Robust Performance on OOD Tasks.}
CrossHA achieves the best success rate on \textit{Craft Items} (78.8\%) and strong performance on \textit{Mine Blocks} (40.0\%) and \textit{All Tasks} (49.1\%), outperforming other RL baselines. CrossHA (w/o STRL) remains competitive but consistently underperforms the full model, especially on \textit{Craft Items} (58.0\% vs. 78.8\%) and \textit{All Tasks} (39.7\% vs. 49.1\%), showing that STRL improves generalization. In contrast, single-space agents show larger variance and weaker transfer outside their specialized action spaces.

\paragraph{Analysis of the Generalization Gap (ID vs. OOD).}
Baselines such as RawHA-RL and MotionHA-RL achieve very high ID performance, e.g., 96.2\% for RawHA-RL on Craft Items, but drop sharply on OOD tasks. In contrast, CrossHA performs strongly on ID tasks, with 83.7\% on Craft Items and 68.8\% overall, while maintaining a smaller generalization gap. This suggests that dynamic action-space selection helps mitigate overfitting and better transfers learned environment knowledge to unseen tasks.

The benefits of RL fine-tuning, especially when combined with heterogeneous action spaces, far outweigh the risks of overfitting. 

\section{Conclusion}

In this work, we present CrossHA, a unified agentic model that integrates heterogeneous action spaces from high-level scripts to low-level atomic commands within a single policy. Through Single- and Multi-Turn GRPO, CrossHA learns to dynamically select the most suitable action space at each step.

Extensive evaluations on over 800 Minecraft tasks demonstrate that CrossHA achieves state-of-the-art performance, significantly outperforming baselines with fixed action spaces. These results highlight the importance of treating action-space selection as a learnable component for improving performance and generalization.

Future work includes improving the efficiency of multi-turn RL and extending the framework to real-world robotics settings, where additional challenges such as safety and latency arise.


\newpage





% ...


% ...

\bibliographystyle{abbrvnat}
\nobibliography*
\bibliography{main}

\newpage

\appendix
\clearpage

\section{Environment Details}
\label{app:environment_detail}

In this work, we utilize Minecraft as the primary testbed for evaluating our agent's capabilities in multi-task reinforcement learning following large-scale pretraining. Minecraft provides a rich, dynamic environment that necessitates both embodied interactions (e.g., navigation, combat) and complex GUI-based operations (e.g., crafting, inventory management). This duality makes it an ideal benchmark for testing generalist agents capable of mastering diverse, open-ended tasks ranging from low-level motor control to high-level strategic planning.

\subsection{Environment Configuration}
Minecraft is an open-world sandbox game where players interact with and modify a procedurally generated 3D world. The game offers a vast array of tasks that resemble real-world challenges. Consequently, agents must navigate complex terrains, manage sparse rewards, and engage in long-horizon planning.
Additionally, the game's popularity provides a wealth of potential training data, including gameplay videos and textual tutorials, making it invaluable for research into large-scale learning from unstructured data.

\begin{figure*}
    \centering
    \includegraphics[width=1.0\linewidth]{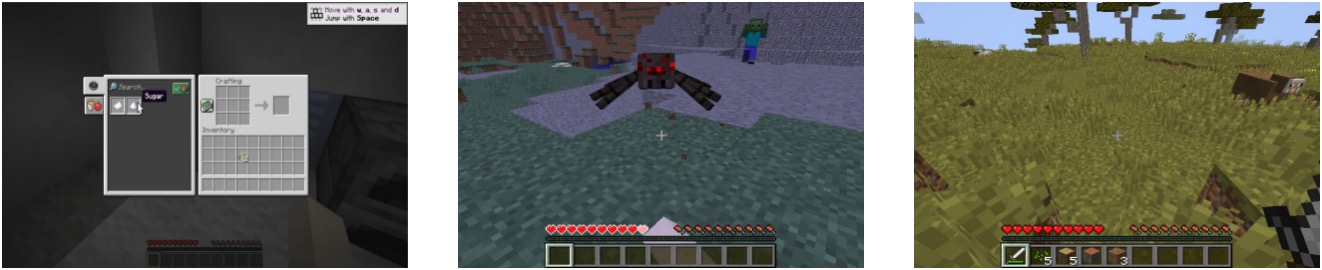}
    \caption{Representative observations in Minecraft, consisting of \(640 \times 360\) RGB images. The agent must handle diverse visual contexts, including embodied first-person views (left) and GUI-based interfaces (right) for tasks like crafting and inventory management.}
    \label{fig:obs}
\end{figure*}

Our experimental setup follows the Variable Pretraining Task (VPT) protocol~\citep{vpt}, where agents interact with the game in a manner identical to human players.
Specifically, the \textbf{observation space} consists exclusively of raw RGB screenshots as visual input, with a resolution of \(640 \times 360\), at a frequency of 20 Hz. No high-level or privileged state information (such as voxel grids or coordinate data) is provided during evaluation.
Regarding \textbf{interaction modes}, the environment presents a dual challenge: while most observations consist of embodied first-person views, tasks such as crafting and smelting require the agent to operate within distinct GUI interfaces. As illustrated in \autoref{fig:obs}, the agent must seamlessly switch between handling continuous embodied interactions and discrete GUI-based operations.

\subsection{Raw Action Space}
The raw action space is designed to align with human interfaces, utilizing the native mouse and keyboard controls provided by MineRL v0.4 (Minecraft version 1.16.5)~\citep{minerl}. A summary of the action mapping is provided in \autoref{tab:action_space}.

\paragraph{Mouse Control.} Mouse displacements are discretized into 1800 bins, and their semantic meaning is context-dependent. In \textbf{embodied mode}, these displacements control the camera orientation via pitch (\(\Delta y\)) and yaw (\(\Delta x\)) adjustments; in \textbf{GUI mode}, they correspond to 2D cursor movements on the screen. Mouse clicks are encoded as dedicated tokens representing left, right, and middle button presses.

\paragraph{Keyboard Control} Keyboard actions are treated as unique tokens, covering alphabetic characters (e.g., `W', `A', `S', `D' for movement), numeric digits (for hotbar selection), and special keys (e.g., \texttt{Shift}, \texttt{Space}, \texttt{Esc}).

\subsection{Challenges for Reinforcement Learning}
Minecraft presents several unique challenges that test the robustness and generalization capabilities of RL agents:

\paragraph{High Information Density}
The environment is visually complex, featuring diverse textures, lighting conditions, and objects. Agents must efficiently process high-dimensional sensory inputs to identify relevant cues while filtering out irrelevant background noise. Specific challenges include distinguishing block types for crafting or spotting hostile mobs in low-light conditions. This requires strong visual representation learning and attention mechanisms.

\paragraph{Task Variety and Generalization}
The open-ended nature of Minecraft offers a vast spectrum of tasks, ranging from primitive resource gathering to complex architectural construction. Agents must learn to generalize skills across these tasks. For instance, the motor skills required to chop a tree can be adapted to combat, while the logic used for crafting a wooden pickaxe serves as a foundational step for crafting diamond tools. This variety forces the agent to acquire adaptable, transferable strategies rather than memorizing fixed sequences.

\paragraph{Reward Sparsity}
Unlike arcade games with frequent score updates, Minecraft rewards are often extremely sparse and delayed. For example, crafting a high-level item (e.g., a diamond sword) requires a long sequence of prerequisite actions: mining wood, crafting a crafting table, mining stone, smelting iron, and finding diamonds. The agent typically receives no feedback until the final goal is achieved. This necessitates efficient exploration strategies and the ability to reason over long horizons.

\paragraph{Dynamic Environment}
The game features a dynamic world with day-night cycles, changing weather, and autonomous entities (mobs). An agent must adapt its decision-making to these variables—for example, seeking shelter at night to avoid monsters or navigating slippery terrain during rain. This dynamic nature ensures that agents are tested on their ability to adapt to non-stationary environments.

\begin{table}[h]
\centering
\caption{Summary of the human-aligned raw action space used by our agent. The agent interacts via standard keyboard and mouse inputs, identical to a human player.}
\label{tab:action_space}
\resizebox{\linewidth}{!}{
\renewcommand\arraystretch{1.1}
\begin{tabular}{@{}ccl@{}}
\toprule
\textbf{Action Type}  & \textbf{Human Input} & \textbf{Description} \\ \midrule
\multirow{7}{*}{Movement} & W                 & Move forward. \\
                          & S                 & Move backward. \\
                          & A                 & Strafe left. \\
                          & D                 & Strafe right. \\
                          & Space             & Jump / Swim up. \\
                          & Left Shift        & Sneak (prevents falling off edges). \\
                          & Left Ctrl         & Sprint (increases movement speed). \\ \midrule
\multirow{2}{*}{Interaction} & Left Mouse     & Break blocks (hold) / Attack entities (click). \\
                             & Right Mouse    & Place blocks / Interact with items / Open GUIs. \\ \midrule
Inventory                 & Keys 1–9          & Select corresponding hotbar slot. \\ \midrule
\multirow{2}{*}{Camera}   & Mouse X           & Yaw: Horizontal rotation (\(-180^\circ\) to \(180^\circ\)). \\
                          & Mouse Y           & Pitch: Vertical rotation (\(-180^\circ\) to \(180^\circ\)). \\
\bottomrule
\end{tabular}
}
\end{table}

\section{Action Spaces}
\label{app:action_spaces}

In the context of Hierarchical Agents (HA), the design of the action space is pivotal, effectively dictating the agent's granularity of interaction with the environment. Existing approaches utilize varying levels of abstraction, spanning from atomic environmental controls to high-level semantic planning. The controllers associated with these spaces range in complexity from simple heuristic scripts to sophisticated Large Language Models (LLMs). Typically, as the level of abstraction increases, the reliability of execution becomes harder to guarantee, necessitating more robust decoders to bridge the gap between intent and execution.

\vspace{0.5em}
\noindent \textbf{Raw Actions} represent the lowest level of the hierarchy. They directly link the agent to the environment's native interface by mimicking basic input devices, such as keyboards and mice~\citep{zhong2025survey, openvla}.

\vspace{0.5em}
\noindent \textbf{Language Skills} operate at a high level of abstraction, where the action space comprises goal-oriented commands expressed in natural language. These semantic instructions are interpreted by the agent's policy, which decodes them into sequences of lower-level primitives conditioned on the current visual observation~\citep{deps, palme}.

\vspace{0.5em}
\noindent \textbf{Motion Actions} serve as an intermediate abstraction, focusing on temporally extended, object-agnostic movement primitives. By encapsulating complex navigational or manipulation behaviors without binding them to specific object instances, they offer a more flexible representation compared to step-by-step atomic actions~\citep{steve1, rth}.

\vspace{0.5em}
\noindent \textbf{Grounding Actions} further enhance abstraction by explicitly incorporating spatial constraints. This paradigm enables the agent to target specific objects based on their coordinates in visual space. By decoupling the semantic intent of an action from its spatial execution parameters, grounding actions significantly improve generalization across diverse visual scenarios~\citep{dexgraspvla, rocket1, lee2025molmoact}.

\vspace{0.5em}
\noindent \textbf{Latent Actions} are derived via self-supervised learning rather than manual definition. Typically encoded into continuous embeddings and subsequently discretized into tokens, these actions allow the agent to operate within a learned abstract space. This approach eliminates the need for human-engineered heuristics, facilitating the acquisition of complex policies directly from large-scale offline data~\citep{vq, deng2025open, yuan2024pre, BehaviorTransformer}.

Collectively, these diverse action spaces contribute to the hierarchical structure of modern agents, providing a scalable and flexible framework for managing tasks of increasing complexity.
\section{RL Algorithms}
\label{app:rl_algorithm}

As detailed in \autoref{sec:method}, our training framework incorporates two reinforcement learning phases: Single-Turn RL (STRL) and Multi-Turn RL (MTRL). Both phases utilize Group Relative Policy Optimization (GRPO) as the underlying optimization algorithm.

\subsection{Single-Turn GRPO}
Standard GRPO optimizes the policy by leveraging group-level statistics. The optimization objective follows the formulation defined in \autoref{eq:st_grpo} in \autoref{sec:method}. 

Unlike PPO, GRPO calculates the advantage $\hat{A}_{i}$ directly from the sampled group outputs without a critic network:
\begin{equation}
\hat{A}_{i} = \frac{r_i - \text{mean}(\{r_1, \dots, r_G\})}{\text{std}(\{r_1, \dots, r_G\}) + \delta}
\end{equation}
where $r_i$ is the reward for the $i$-th output, and $\delta$ (e.g., $10^{-8}$) is a small constant added for numerical stability.

The inclusion of the KL-divergence penalty term $\beta D_{KL}$ acts as a regularizer, ensuring the updated policy does not deviate significantly from the reference model, thereby preventing reward hacking and maintaining linguistic coherence.

\subsection{Multi-Turn GRPO}
In the multi-turn setting, we adapt the single-turn formulation to handle sequential interactions. We decompose each trajectory $\tau$ into a set of state–response pairs and assign the final trajectory-level reward \(r(\tau)\) to all intermediate responses.

Let $\mathcal{T}$ denote the trajectory space. The Multi-Turn GRPO objective is defined as:
\begin{align}
J_{\text{MT}}(\theta) &= \mathbb{E}_{\{\tau_i\}_{i=1}^G \sim \pi_{\text{old}}} \Bigg\{ \frac{1}{G} \sum_{i=1}^{G} \frac{1}{L_i}\sum_{j=1}^{L_i} \frac{1}{|o_{i,j}|}\sum_{t=1}^{|o_{i,j}|} \Bigg[ \nonumber \\
& \min \Bigg( \rho_{i,j,t} \hat{A}_{i}, \text{clip} \Big( \rho_{i,j,t}, 1 - \epsilon, 1 + \epsilon \Big) \hat{A}_{i} \Bigg) \nonumber \\
& - \beta D_{\text{KL}} [ \pi_{\theta}(\cdot|s_{i,j}) || \pi_{\text{ref}}(\cdot|s_{i,j}) ] \Bigg] \Bigg\}
\label{eq:mt_grpo}
\end{align}
where $L_i$ is the number of turns in trajectory $i$, and the advantage $\hat{A}_{i}$ is computed using the group-relative statistics of the trajectory rewards $r(\tau_i)$. This allows the sparse, episodic reward to guide the optimization of every step in the trajectory.

\begin{table}[h]
\centering
\caption{Hyperparameter settings across different training stages.}
\label{tab:training_setup}
\resizebox{\linewidth}{!}{%
\renewcommand\arraystretch{1.2}
\begin{tabular}{@{}l c c c@{}}
\toprule
\textbf{Hyperparameter} & \textbf{SFT} & \textbf{STRL} & \textbf{MTRL} \\ \midrule
Total Training Tokens & 0.07B & 0.14B & 1.41B \\
Training Samples (Images) & 12K & 24K & 1.3M \\
Trainable Components & Full & Full & Full \\
Global Batch Size & 512 & 1272 & 128 \\
Optimizer & AdamW & AdamW & AdamW \\
LR Warmup Steps & 15 & 0 & 0 \\
Maximum Learning Rate & $8.0\times10^{-6}$ & $1.0\times10^{-6}$ & $5.0\times10^{-6}$ \\
Minimum Learning Rate & $1.0\times10^{-6}$ & $1.0\times10^{-6}$ & $5.0\times10^{-6}$ \\ 
Group Size ($G$) & - & 4 & 4 \\
KL Coefficient ($\beta$) & - & 0.01 & 0.01 \\
Clipping Parameter ($\epsilon$) & - & 0.2 & 0.2 \\
\bottomrule
\end{tabular}%
}
\end{table}

\section{Training Details}
\label{app:training_details}

\subsection{Base Model}
To enable robust multi-modal reasoning and high-level planning, our system is built upon a pre-trained Vision-Language Model (VLM). Specifically, we initialize our system with the open-source \textbf{OpenHA} model~\cite{openha}, which is derived from Qwen2-VL-7B~\cite{qwen2vl}. We selected this model for its demonstrated proficiency in Minecraft-specific visual grounding and its strong instruction-following capabilities within the domain.

\subsection{Training Implementation}
\label{app:training_setup}
As detailed in \autoref{sec:method}, our training pipeline comprises three progressive stages: Cold-Start Supervised Fine-Tuning (SFT), Single-Turn Reinforcement Learning (STRL), and Multi-Turn Reinforcement Learning (MTRL).
To ensure experimental consistency and reproducibility, all models were trained on a cluster of 8 NVIDIA A800-SXM4-80GB GPUs, using Python 3.10 and CUDA 12.6.

The specific hyperparameter configurations for each stage are summarized in \autoref{tab:training_setup}.
We maintain consistent hyperparameter settings for the SFT process across both the initial Stage 1 and the cold-start phases to isolate the impact of training data and objectives

\begin{figure}[!t]
    \centering
    \includegraphics[width=1.0\linewidth]{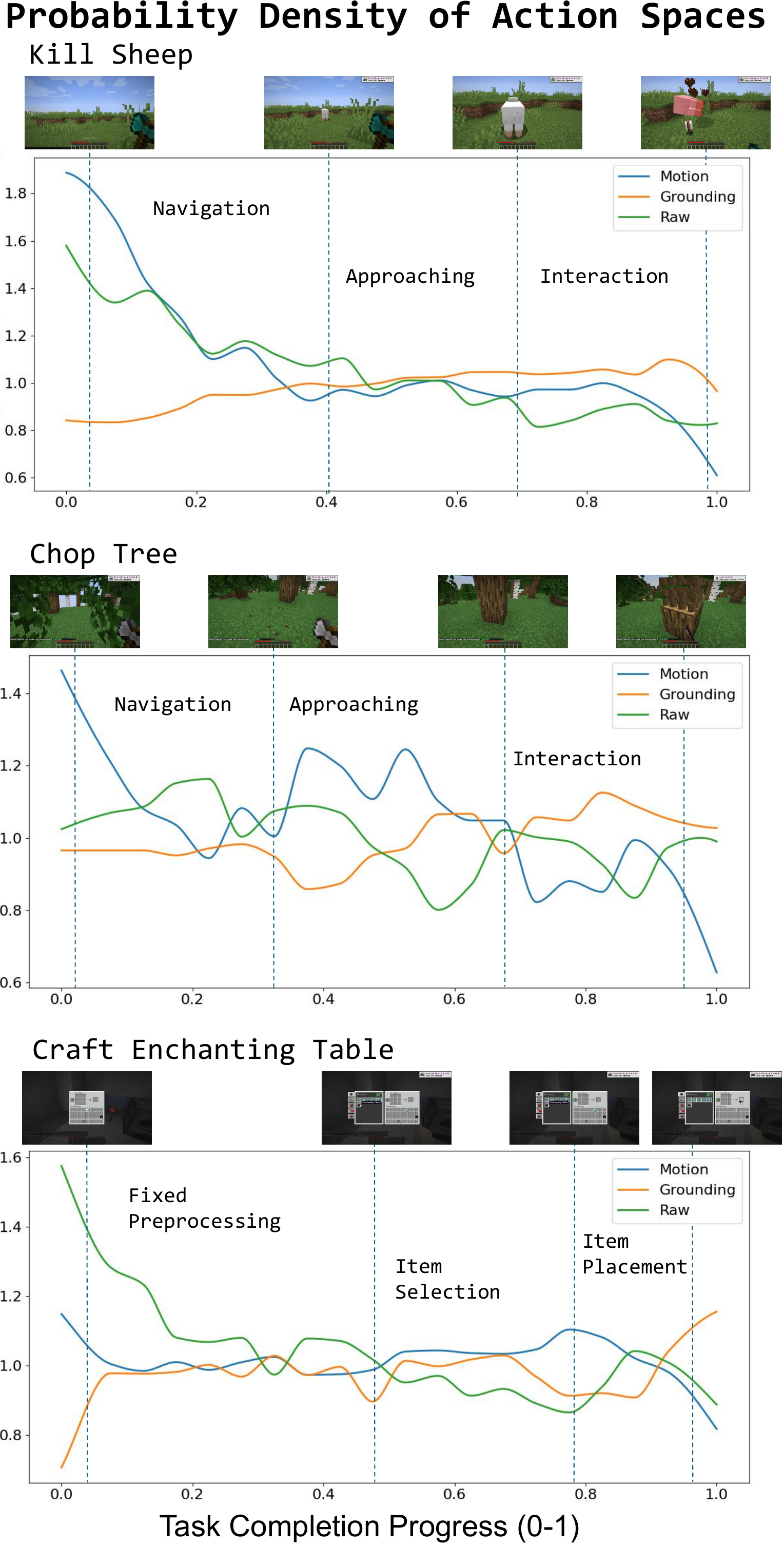}
    \caption{\textbf{Case Study: Action distribution during the \texttt{Kill Sheep}, \texttt{Chop Tree} and \texttt{Craft Enchanting} task.} The density curves of each tasks, aggregated over 20 episodes, of different action spaces (Motion, Grounding, Raw) across different task phases. The dynamic shifts in distribution demonstrate the model's in-context adaptive strategy.}
    \label{fig:example_rollout}
\end{figure}

\begin{figure*}[t]
    \centering
    \includegraphics[width=0.95\textwidth]{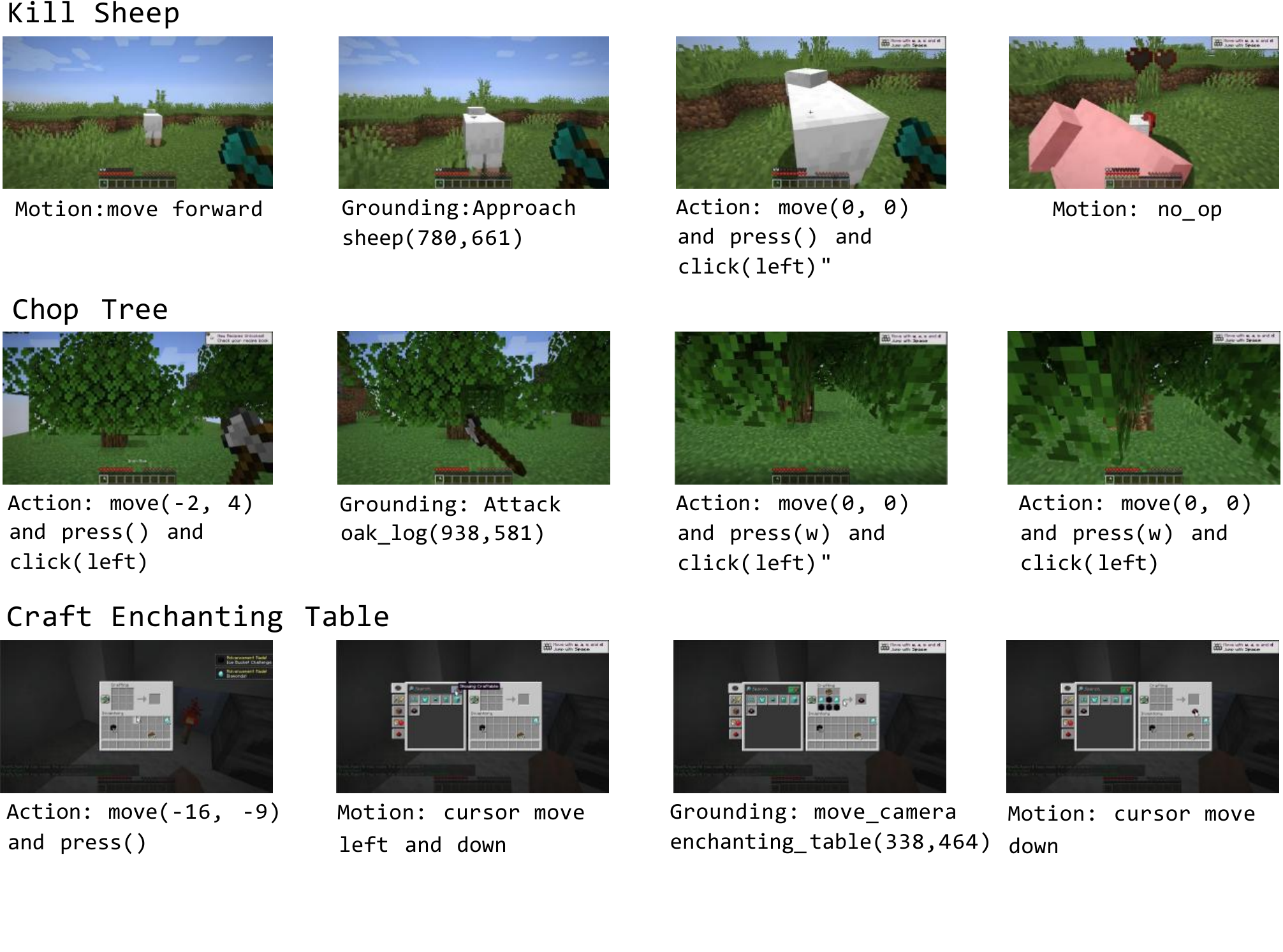}
    \caption{\textbf{Example Rollouts of \texttt{Kill Sheep}, \texttt{Chop Tree}, and \texttt{Craft Enchanting}.}
    The prefixes ``Motion:'', ``Grounding:'', and ``Action:'' denote actions from the Motion Space, Grounding Space, and Raw Space, respectively.}
    \label{fig:case_study}
\end{figure*}

\section{Case Studies}

\label{app:case_study}
To qualitatively demonstrate the efficacy of dynamic action space selection, we analyze CrossAgent's behavior across three representative tasks: \texttt{Kill Sheep} (Kill Entities), \texttt{Chop Tree} (Mine Blocks), and \texttt{Craft Enchanting Table} (Craft Items). These cases illustrate how the agent adapts its interface strategy to meet the distinct demands of different task phases. \autoref{fig:case_study} illustrates the probability density curves of different action spaces selected by CrossAgent with respect to task completion progress.

\paragraph{Kill Sheep}
As illustrated in the \texttt{Kill Sheep} task proceeds through three distinct stages:
\textbf{1) Navigation}: Rotating the camera and traversing the terrain to locate a target.
\textbf{2) Approaching}: Locking focus on the sheep and closing the distance.
\textbf{3) Interaction}: Executing attacks to deal damage.
During the \textbf{Navigation Stage}, CrossAgent predominantly employs the coarse yet efficient Motion Action Space to scan large areas rapidly. Once the sheep enters the field of view, the agent transitions to the \textbf{Approaching Stage}, characterized by a balanced mix of action spaces: Motion Actions facilitate fast movement, Grounding Actions provide continuous visual tracking cues, and Raw Actions enable fine-grained adjustments to navigate around obstacles. Finally, in the \textbf{Interaction Stage}, sustained Grounding Actions ensure precise targeting of the moving entity, while Raw Actions are leveraged to execute high-frequency attack commands efficiently.
This progression demonstrates CrossAgent's ability to switch contextually: prioritizing search efficiency initially, then blending modalities for pursuit, and finally optimizing for targeting precision.

\paragraph{Chop Tree}
The \texttt{Chop Tree} task shares the Navigation-Approaching-Interaction structure with \texttt{Kill Sheep} but exhibits distinct characteristics. Trees are static targets, and the agent typically spawns in forested areas where targets are abundant. Consequently, the task requires less precise tracking than hunting moving entities. Motion Actions dominate the navigation and approaching phases. Unlike the \texttt{Kill Sheep} task, where the agent must constantly adjust its bearing to chase a target, the static nature of trees allows the agent to rely heavily on the simple `Walk Forward' primitive within the Motion Space. In the interaction phase, Grounding Actions are engaged to accurately target the specific wood block, ensuring successful harvesting.

\paragraph{Craft Enchanting Table}
Crafting tasks involve complex, multi-step GUI interactions. We categorize the workflow into three phases:
\textbf{1) Fixed Preprocess}: Opening the recipe book and toggling the `Only Show Craftable Items' filter. Since these steps are invariant across tasks, Raw Actions dominate this phase, leveraging their precision and stability to execute rote "muscle memory" sequences efficiently.
\textbf{2) Item Selection}: Locating and selecting the target item within the grid. Here, the identification capability of the Grounding Space becomes crucial for visually recognizing the correct icon. Simultaneously, the Motion Space aids in smooth cursor traversal.
\textbf{3) Item Placement}: Retrieving the crafted item from the result slot. The Grounding Space is again pivotal, helping the model accurately coordinate the cursor between the result slot and the inventory, mitigating the risk of mis-clicks.

\autoref{fig:example_rollout} further illustrates some key frames and corresponding outputs from the CrossAgent rollout.  These examples demonstrate CrossAgent’s ability to adaptively select the suitable action space in context, leading to more precise and efficient task execution.

The qualitative rollouts corroborate our statistical findings, offering a concrete window into the agent's decision-making process. The generated trajectories reveal that CrossAgent's switching mechanism is semantically coherent rather than stochastic. For instance, the seamless transition from high-level \textit{Motion} primitives for efficient exploration to precise \textit{Grounding} or \textit{Raw} commands for interaction closely mirrors human behavioral patterns. This qualitative evidence, combined with the quantitative performance metrics, strongly validates that CrossAgent has successfully internalized a hierarchical understanding of open-world tasks. It demonstrates that the model does not merely memorize trajectories but learns to autonomously decompose complex objectives into optimal sequences of heterogeneous actions, thereby achieving both robustness and efficiency.

\noindent

\end{document}